\documentclass[conference]{IEEEtran}
\usepackage{amsmath,amsfonts}
\usepackage{algorithmic}
\usepackage{algorithm}
\usepackage[numbers]{natbib}
\usepackage{array}
\usepackage[caption=false,font=normalsize,labelfont=sf,textfont=sf]{subfig}
\usepackage{textcomp}
\usepackage{stfloats}
\usepackage{url}
\usepackage{verbatim}
\usepackage{graphicx}
\usepackage{subfig}
\usepackage{svg}
\usepackage{cite}
\usepackage[separate-uncertainty=true]{siunitx}
\newcommand{\minisection}[1]{\vspace{0.025in} \noindent {\bf #1} \ }

\begin{document}

\title{Reinforcement Learning for High-Level Strategic Control in Tower Defense Games}

\author{
\IEEEauthorblockN{Joakim Bergdahl}
\IEEEauthorblockA{\textit{SEED - Electronic Arts (EA)} \\
Stockholm, Sweden \\
jbergdahl@ea.com}
\and
\IEEEauthorblockN{Alessandro Sestini}
\IEEEauthorblockA{\textit{SEED - Electronic Arts (EA)} \\
Stockholm, Sweden \\
asestini@ea.com}
\and
\IEEEauthorblockN{Linus Gisslén}
\IEEEauthorblockA{\textit{SEED - Electronic Arts (EA)} \\
Stockholm, Sweden \\
lgisslen@ea.com}

}


\IEEEoverridecommandlockouts
\IEEEpubid{\makebox[\columnwidth]{ 979-8-3503-5067-8/24/\$31.00~\copyright2024 IEEE \hfill} 
\hspace{\columnsep}\makebox[\columnwidth]{ }}

\maketitle

\IEEEpubidadjcol

\begin{abstract}

In strategy games, one of the most important aspects of game design is maintaining a sense of challenge for players. Many mobile titles feature quick gameplay loops that allow players to progress steadily, requiring an abundance of levels and puzzles to prevent them from reaching the end too quickly. As with any content creation, testing and validation are essential to ensure engaging gameplay mechanics, enjoyable game assets, and playable levels. In this paper, we propose an automated approach that can be leveraged for gameplay testing and validation that combines traditional scripted methods with reinforcement learning, reaping the benefits of both approaches while adapting to new situations similarly to how a human player would. We test our solution on a popular tower defense game, \textit{Plants vs. Zombies}. The results show that combining a learned approach, such as reinforcement learning, with a scripted AI produces a higher-performing and more robust agent than using only heuristic AI, achieving a 57.12\% success rate compared to 47.95\% in a set of 40 levels. Moreover, the results demonstrate the difficulty of training a general agent for this type of puzzle-like game.
\end{abstract}


\section{Introduction}
\label{sec:intro}
The mobile games market is constantly growing and had, as of 2021, the biggest share (50\%) of the total gaming market (\$176B) \citep{bernes2022comparison}. Modern mobile games can be nearly as content-rich and engaging as AAA computer or console games, which puts new demands on the developers of this type of games that might not have existed a couple of years ago.
Many mobile titles are built around quick gameplay loops where the player is progressing through the game at a steady pace. As with any content creation, it needs to be tested and validated. Gameplay mechanics must be engaging, game assets must be enjoyable and levels playable.

This requires an abundance of levels and puzzles to stop the player from reaching the end of the game too quickly. Moreover, many mobile games are live-service titles: long-lasting games that, once published, developers continue to add mechanics, assets, and especially levels. Everything that is added after initial distribution, must be thoroughly tested. One example where this is particularly true is the tower defense genre. Tower defense games are a sub-genre of strategy games, in which the player must defend their territory from waves of enemies, typically by placing defensive units to counter the attacking forces. Each level presents a puzzle that the player must solve by identifying the right strategies to follow and positioning the appropriate units optimally at each step. Tower defense games are popular in the mobile gaming landscape, with notable examples including \textit{Plants vs. Zombies}, \textit{Bloons TD}, and \textit{Clash Royale}.
 
One of the big challenges with creating engaging games of this type is the tuning of the level of difficulty. This is especially true for more casual games, such as those in mobile platforms. Often, game designers have just minutes to catch the player's interest and without a well-balanced challenge it could be hard to keep the player from progressing farther into the game. In order for a mobile game to be engaging, it has to have a carefully measured level of difficulty. Too difficult and the gameplay experience becomes frustrating. Too easy and the experience simply becomes unrewarding.

There are many ways of testing the difficulty level of a game. One of the more commonly used methods is to leverage the game's soft-launch phase to gather data on playthroughs and their respective players. However, it is not an ideal way to use live player data to fine-tune the game: it might result in the loss of many potential long-term players that otherwise would have continued to play the game. Another option is to use automated scripts to play the game and in that way estimate the difficulty. A scripted solution is seldom powerful enough to handle new levels and features that are constantly added both in production, and post-production for a live-service games. 

Here, we propose a method where we bring together the benefits from human play-testers (adaptive, learning) and the scripted solution (fast, scalable, cheap). In this work, we propose a hybrid approach that combines reinforcement learning (RL) and scripted heuristic AI (HAI). By leveraging the best qualities of the two approaches, one can create a testing AI that is efficient and, at the same time, high-performing, with a higher level of adaptability to changes compared to a HAI-only method. The approach is fairly simple, yet quite powerful: we train the RL agent as a high-level decision-maker, which selects the low-level action the HAI system should perform, deferring the responsibility of how the action is executed to the latter. 

We test our approach in a popular and previously mentioned tower defense game: \textit{Plants vs. Zombies}. This game has all the main characteristics of this genre: a live-service puzzle game played by millions of players, with a simple design but complex gameplay mechanics that require thoughtful strategies. We show how the combination of RL and HAI when trained for each single level performs better than using only the latter. However, we show how this type of puzzle game is hardly generalizable, with each level needing a particular high-level strategy. 

\section{Related Work}
In this section we review the most relevant literature to our contributions.  

\subsection{Learning Agents in Games}
RL has garnered great interest from the video game community the last few years. Major developments in optimization algorithms and model architectures has lead to impressive results in complex games such as \textit{Star Craft II}, \textit{Dota 2}, and \textit{Gran Turismo 7} \citep{alphastar, openai2019dota, gtsophy}. However, these works focused primarily on training the best agent to replace human players, mainly leveraging RL. More recent works explore different methods for training agents in various use cases such as creating believable and human-like game AI \citep{navigates, counterstrike, voyager}. Few approaches have used self-learning agents in mobile games, especially puzzle-like games such as tower defense. Notable examples include: the work by \citet{rltower}, where the authors proposed an RL pipeline for a tower defense player agent; the work by \citet{kristensen2021}, which will be examined in a later section; and CandyRL \citep{candyrl}, which employs an approach similar to ours for playing Match-3 games.

As mentioned in Section~\ref{sec:intro}, automated playtesting through learning agents has gained interest in both research and industry community. However, when dealing with a game in development, training agents with RL seems to be the most feasible solution, particularly during the playtesting phase. A game in development lacks pre-collected datasets, and creating data from expert demonstrations can be time-consuming. Furthermore, it is crucial to be fast and efficient in order to stay in sync with game developers.


\subsection{Automated Playtesting}

Many works have employed machine learning for automated gameplay testing~\citep{towardstest}.  \citet{mugrai2019automated} showed that mimicked human behavior can be used to achieve more meaningful gameplay testing. \citet{kiwi} utilized a data-driven technique that allows designers to efficiently train game testing agents, explaining why learning agents are essential for playtesting. \citet{augmenting} proposed a study on the usefulness of using RL policies to playtest levels against scripted agents. \citet{ccpt} proposed the curiosity-conditioned proximal trajectories algorithm, which tests complex 3D game scenes using a combination of IL, RL, and curiosity-driven exploration. Finally, \citet{xboxplaytesting} proposed an agent that primarily relies on pixel-based state observations while exploring the environment, conditioned on a user's preference specified by demonstration trajectories. 

However, as we will discuss throughout the paper, if one were to employ a RL-only solution for grid-based tower defense games like \textit{Plants vs. Zombies}, the vast number of available actions would make credit assignment particularly difficult. For this reason, we opted for a hybrid approach that combines RL with the pre-existing HAI. Similar approaches have been explored by \citet{candyrl}, \citet{match3_playtesting}, and \citet{starcraft_combining}, but for different types of games such as Match-3 and real-time strategy games.

\subsection{Difficulty Evaluation}
Estimating and evaluating game difficulty is an active topic in video game related research, as it is an important component in good game design. Classically, difficulty has been evaluated and tuned following extensive playtesting by human play-testers, an endeavour that is both time-consuming and expensive. Statistical modeling of difficulty has shown promise in puzzle-style games where the number of moves is limited, successfully predicting the impact on level difficulty when modifying the number of moves allowed \citep{kristensen2021}. Other researchers have explored using machine learning in a tower defense game to adjust the actual difficulty dynamically in an otherwise static system, managing how enemy waves are spawned \citep{massoudi2013}. Research applied to the tower defense game \textit{Kingdom Rush: Frontiers} approaches the problem of difficulty assignment through procedural content generation by using flat Monte Carlo search to verify that generated levels are playable, but also that they have a balanced difficulty \citep{liu2019}. All the aforementioned approaches require data collected from well-playing players, that could be difficult to source. As already described in Section~\ref{sec:intro}, this paper proposes an approach that combines the benefits from human testers but also from heuristic AI. Developers can leverage on this approach to collect high-quality data for running statistical analyses for difficulty evaluation and validation.


\section{Environment}
\begin{figure}
    \centering
    \includegraphics[width=\columnwidth]{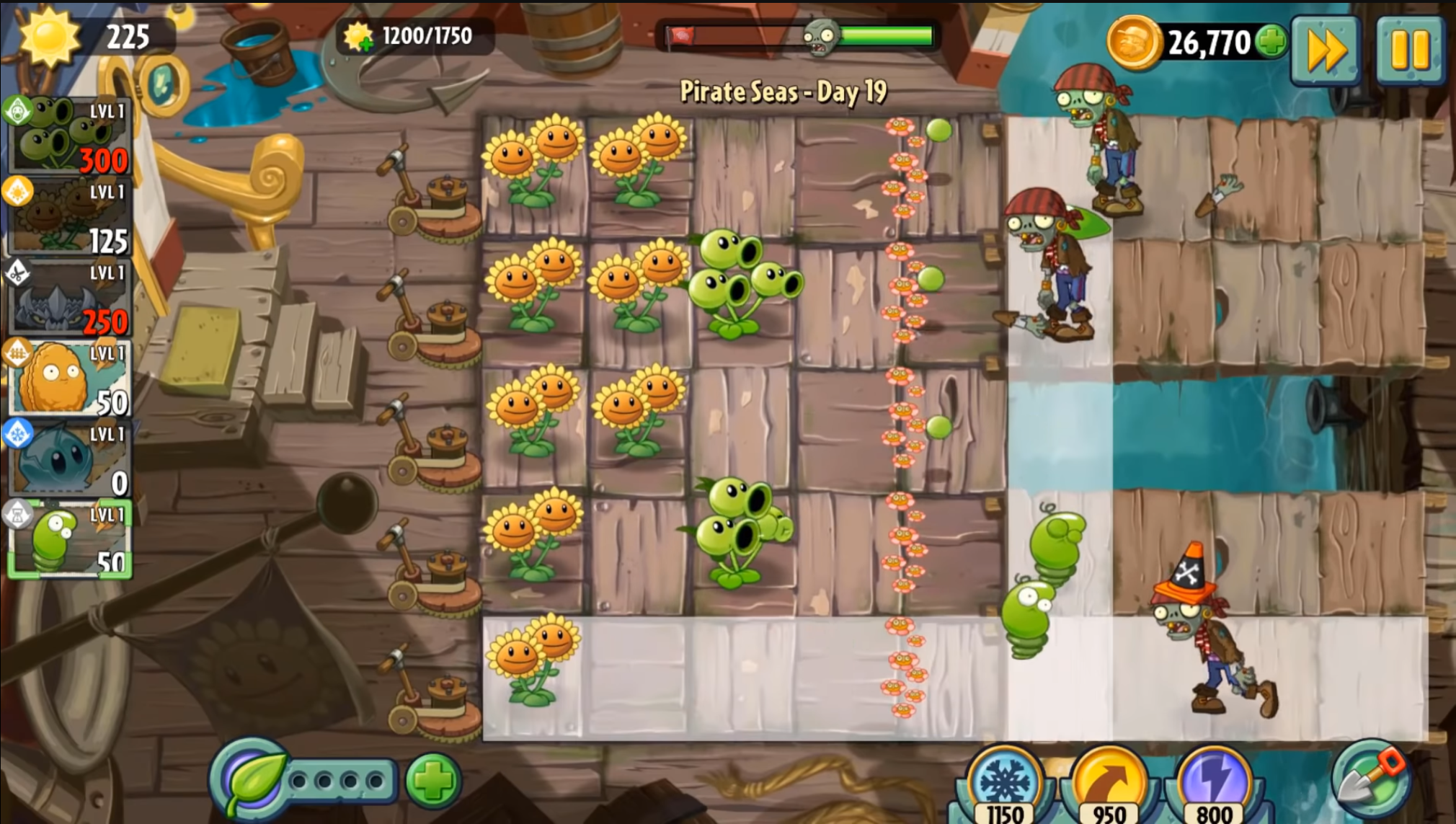}
    \caption{Screenshot of \textit{Plants vs. Zombies 2} gameplay. Enemy zombies attack the player from the right, who has to defend their home base line on the leftmost side of the playarea. If an enemy crosses this line, the player loses. The player is free to place units in any available, unoccupied cell of the game board from the loadout visible on the left. When a unit is greyed out, it is locked behind a cool-down timer. Each unit's sun token cost is displayed in the units lower right corner.}
    \label{fig:pvz}
\end{figure}

In this section we dive deeper into the \textit{Plants vs. Zombies} (PvZ) game. Figure~\ref{fig:pvz} shows a screenshot of the game.

\subsection{Game Description}
In PvZ, the player is tasked to use a set of plants to defend themselves against incoming waves of enemy zombies. What makes PvZ unique is its grid-based nature where the player is limited to place their plants in predefined grids with different layouts of at most five rows, or \textit{lanes}, by nine columns. Similar to other tower defense titles, the complexity of PvZ comes from long-term planning, resource management and the individual abilities of the plant units at the player's disposal.
In each level, PvZ presents all the common tower defense elements:
    1) a base that must be defended from attacking enemies;
    2) one or more waves of enemies, the diversity and strength of each depending on the particular level and its difficulty; and
    3) units with varying defense/offense characteristics required to defend the base against attacking enemies.

At the start of a round, the player is given a predefined set of plant units, called a \textit{loadout}, they can use in the level. As the level starts, the player is tasked to prepare for the first wave of enemies by placing plant units in unoccupied grid cells. Doing so, the player needs to spend \textit{sun tokens} which are either generated passively in the game-world at a slow rate or through specific plant units that produce them. Each unit incurs a sun token cost, and more capable units tend to be more expensive, demanding planning and prioritization. Importantly, key units will help the player defeat enemies or hinder their advancement. Whenever any unit from the loadout has been placed, this unit will not be usable until a corresponding cool-down timer runs out. This further limits the rate with which players can perform actions, promoting strategy instead. As the enemy waves start spawning, this loop of resource collection and preparation repeats. The termination or game-over criteria for PvZ is fulfilled in two conditions: 1) the player successfully defeats all enemy units and wins; 2) one or more enemy units reach the left-most side of the play field and the player loses.

\subsection{Heuristic AI and Automated Playtesting}
\label{sec:hai}
When experiments started, the PvZ game had employed an HAI system capable of playing the game to check for stability issues and performance drops. HAI replaces the player and it is composed of two parts: one for computing priority values over a set of strategies, and another for running the strategy with the highest priority. The current implementation of HAI contains 4 of these strategies and we will go into detail about each of them later in Section~\ref{sec:strategies}. At each step, the HAI takes the current state of the game as input and outputs a priority value for each low-level action. The system then executes the strategy with the higher priority value. We must note that the way the HAI computes the priority values \emph{remains the same across levels}. If the current state is more or less the same but in two different levels, the priority values will be the same. This is a suboptimal solution because, as we will see with our experiments, some levels require dynamic values to be actually played, and a particular play-style -- e.g., preferring defensive plants over attacking ones -- can work in one level but not in another. Moreover, if developers want to add a new strategy or a new unit, they need to rebalance all the values.

The HAI system is crucial for PvZ developers, as it allows them to test the game at a very high speed. However, for the reasons listed above, HAI is not general, hardly maintainable, and poorly scalable. Thus, we decided to replace the first system with an RL agent.


\section{Method}
In this section, we explain our proposed hybrid method that combines HAI and RL, referred to as hybrid RL (HRL) for the remainder of the paper. First, we describe the 4 strategies used by both our hybrid approach and HAI. Then, we describe the algorithm used to train the HRL agent that replaces the priority calculation described in Section~\ref{sec:hai} and how it is combined with the HAI action execution.

\subsection{High-Level Strategies}
\label{sec:strategies}
The built-in HAI system spans four strategies, capable of covering all gameplay facets of PvZ.

\minisection{Sow Sun.}
Central to all tower defense games is the resource management aspect. In PvZ, the main resource collected by the player are sun tokens. Without sun tokens, the agent will not be able to produce or upgrade any units that is needed to complete the level. The \textit{sow sun} strategy is responsible for placing units available in the loadout that contribute to the production of sun tokens on the game board. An example of this is the \textit{sunflower} unit, that spawns sun tokens at fixed intervals.


\minisection{Attack.}
In order to successfully complete a game level in tower defense, the player needs to defeat all incoming enemy units, usually split up in individual waves. To do so, the player has a set of offensive units at their disposal. These units vary from close range melee units to ranged shooting units. The \textit{attack} strategy of the classical AI system is responsible for using this combined class of units. In this regard PvZ follows pretty much the typical tower defense setup. An exemplary attack unit is the \textit{peashooter}, that shoots projectiles at approaching enemies in the same lane where it is placed.

\minisection{Defensive.}
As resource generation takes time and unit placement incurs a cool-down timer, the player needs to be able to limit enemy advancement. This is done through the \textit{defense} strategy. Units of this type are mainly passive and made to simply block enemies from moving further, with an example being the \textit{walnut}. These units also tend to have a high amount of hit-points to endure damage over time for longer.

\minisection{Prepare.}
A special strategy employed by HAI is the \textit{prepare} one. This strategy controls placement of passive units that generally deal no damage nor survives enemy attacks. Their purpose is merely to allow placement of other units of the aforementioned types. An example is the \textit{lilypad} unit which, when placed on a water grid cell, allows for another unit to be placed on-top of it, opening up the game board. Without the lilypad, the water cell does not allow for unit placement.

\subsection{Algorithm}
\label{sec:algorithm}
We opted for a combination of RL and HAI, motivated by the following reasons:
\begin{itemize}
    \item As previously discussed, HAI is difficult to maintain, does not generalize well to new levels and/or mechanics, and has poor scalability. Furthermore, it exhibits suboptimal performance in some levels. Lastly, as we will see in Section~\ref{sec:difficulty_robustness}, it lacks robustness when faced with varying levels of difficulty. However, after our initial experiments, we observed that the main issue was primarily due to HAI's strategy selection system, rather than its execution of actions;
    \item A fully-fledged RL agent, trained to select one of the units and place it wherever desired, is hardly feasible in this case: each level features a grid of $5 \times 9$ cells and a layout of up to 6 units. This results in an action space composed of $5 \times 9 \times 6 = 270$ actions. This presents an efficiency challenge for the RL agent as exploring the action space would be time-consuming, especially when learning something that HAI already manages. In our context, efficiency is crucial, as training new agents should keep pace with game development. \citet{technical2023gillberg} provide a comprehensive summary of the requirements for applying RL in game development and explain why combining RL with scripted approaches is a good solution in this context.
\end{itemize}
Hence, we combine the strengths of both approaches while minimizing their respective drawbacks: the hybrid RL agent leverages the knowledge of HAI and determines which low-level strategy the HAI should execute, effectively replacing the strategy selection system of the AI.

In this section, we will go through each of the elements of the employed algorithm. For all the experiments we use proximal policy optimization (PPO) as the optimization method, which is a popular actor-critic on-policy RL algorithm \citep{ppo}. For this reason, we first have to define two networks: one for the actor and one for the critic.

\minisection{Model Architecture.}
For the PPO implementation, the two underlying actor and critic models share no weights and are initialized in separation. Both models are composed of two fully connected layers of size $1024$, all with leaky ReLU activation functions. The actor is then followed by an output layer of size $5$ fed through a softmax function to produce an action distribution, corresponding to the five actions covering the four available strategies and an additional no-op action. The complete state space is described in Section~\ref{sec:environment_details}. More information regarding the action space is provided later in this section. The critic network is followed by an output layer of size $1$ with no activation for the value estimate.

\minisection{Action Masking.}\label{sec:action_mask}
Given the resource requirements and cool-down timers restricting gameplay flow, the agent will encounter states where no action will yield an outcome. To avoid these scenarios, action masking is used. At each simulation step, for a given semantic unit type tied to a specific strategy, each unit of said group is checked for. If there is at least one unit of this group that is cheaper than the available sun token resource pool $S$ and is not currently restricted through a cool-down timer, the action corresponding to this group is labeled as available. This is repeated for each group. In the worst case, no action is available and the game simulation is forwarded to the point where the agent again has agency over the environment, and at least one action is available besides the no-op. 

\section{Experimental Setup}
In this section we detail the experimental setup used to evaluate our hybrid approach. All experiments utilizing $5$ different seeds were run over $5$ machines, each with an Intel Xeon Gold 6230 @ 2.10 GHz CPU, 128 GB of RAM and an NVIDIA RTX A6000 GPU with 11 GB of VRAM. On average, a training run for one model completed in 5 hours and each model was trained for 10K episodes.

\begin{table*}[]
\centering
\caption{Performance comparison of success rates, episodic rewards and episode lengths in steps between HRL, HAI and a random agent. All values represent the mean of 100 episodes with 5 seeds. For HRL, we train one agent for each level. HRL outperforms the baselines in most of the levels, achieving both higher success rates and rewards. In the final row, the average of the success rate and the sum of the rewards and steps for each level are reported. Note, the levels are not ordered by complexity and there is one particular level (level 23) where HAI achieves a very low reward.}
\label{tab:main_table}
\scalebox{0.7}{
\begin{tabular}{c|ccc|ccc|ccc}
      & \multicolumn{3}{c|}{Success Rate ($\%$)}       & \multicolumn{3}{c|}{Cumulative Reward ($\mathbb{R}$)} & \multicolumn{3}{c}{Average Number of Steps}\\ 
Level & HAI          & Random                & HRL & HAI          & Random                & HRL  & HAI          & Random                & HRL  \\ \hline

1 & \multicolumn{1}{c|}{$\phantom{0}14.00 \pm \phantom{00}3.90$} & \multicolumn{1}{c|}{$\phantom{00}6.60 \pm \phantom{00}3.01$} & \multicolumn{1}{c|}{$\mathbf{\phantom{0}29.80 \pm \phantom{0}15.66}$} & \multicolumn{1}{c|}{$\mathbf{\phantom{0}11.56 \pm \phantom{00}0.82}$} & \multicolumn{1}{c|}{$\phantom{00}0.63 \pm \phantom{00}0.93$} & \multicolumn{1}{c|}{$\phantom{0}11.40 \pm \phantom{00}5.23$} & \multicolumn{1}{c|}{$\phantom{}351.02 \pm \phantom{00}9.40$} & \multicolumn{1}{c|}{$\phantom{}263.88 \pm \phantom{00}6.92$} & \multicolumn{1}{c}{$\mathbf{\phantom{}404.37 \pm \phantom{0}64.51}$}\\
2 & \multicolumn{1}{c|}{$\phantom{0}45.80 \pm \phantom{00}2.40$} & \multicolumn{1}{c|}{$\phantom{00}5.40 \pm \phantom{00}0.80$} & \multicolumn{1}{c|}{$\mathbf{\phantom{0}67.00 \pm \phantom{00}9.61}$} & \multicolumn{1}{c|}{$\phantom{00}8.41 \pm \phantom{00}0.48$} & \multicolumn{1}{c|}{$\phantom{00}\llap{-}5.96 \pm \phantom{00}0.21$} & \multicolumn{1}{c|}{$\mathbf{\phantom{0}10.72 \pm \phantom{00}1.49}$} & \multicolumn{1}{c|}{$\phantom{}246.57 \pm \phantom{00}3.52$} & \multicolumn{1}{c|}{$\phantom{}143.71 \pm \phantom{00}3.40$} & \multicolumn{1}{c}{$\mathbf{\phantom{}266.55 \pm \phantom{0}19.17}$}\\
\end{tabular}
}
\end{table*}

\subsection{Environment Details}
\label{sec:environment_details}
In this section we detail the following components: observation space, action space, and reward function.

\minisection{Observation Space.}
This information is represented as feature vector observations of size $88$, composed by:

\begin{itemize}
    \item One-hot encodings of the semantic type of each unit in the loadout with masking if the loadout is incomplete. The maximum number of units in the loadout is $6$, and some levels can have less than $6$ units. In total, for this we have  $6 \times 6$ values;
    \item Current cool-down statuses of each loadout unit. Each value is in $[0,1]$ with 0 meaning the unit is available, and for this we have a total of $6$ values;
    \item Amount of currently held sun tokens, only $1$ value;
    \item Average health points over all enemy units for each row, for a total of $5$ values;
    \item The distances to the closest enemy in each row, for a total of $5$ values;
    \item Number of enemies that have advanced past the mid-point of the board per row, for a total of $5$ values;
    \item Number of enemies per row, for a total of $5$ values; and
    \item Count of planted units of each semantic type for each row, for a total of $5 \times 5$ values.
\end{itemize}

\minisection{Action Space and Execution.}
Deferring low-level strategy execution to the classical AI system, the action space for the HRL agent is simple. The agent only needs to produce one out of four available high-level strategies at each decision step as described in Section \ref{sec:strategies}. In addition, a no-op action is introduced to allow for accumulating resources or waiting for more optimal actions. As the game might be in a state where no action can be performed due to lack of sun tokens or all plants being unavailable due to cool-downs, the simulation is progressed until at least one additional action than the no-op is available through the action mask as detailed in Section~\ref{sec:action_mask}.

\minisection{Reward Function.}
Tower defense games require the player to survive and defeat waves of incoming enemies. As such, the agent gets a positive reward of $1.2 \cdot N$ where $N$ is the number of enemies defeated since the last action decision. To prolong play, allowing for more complicated action selection, the agent is also penalized with a fixed value of $r_{Z} = -1/200$ for each enemy that has been allowed to advance since the last action decision. The reasoning behind this is to make use of the defensive strategy more appealing, leading to more placement of blocking plant units. Finally, the agent receives a terminal reward of either $1$ or $-5$ depending on if the level was successfully completed or the agent lost.

\subsection{Levels and Difficulty}
\label{sec:levels_and_difficulty}
For our experiments, we decided to use a set of $40$ already existing levels. Although they do not encompass all the levels in the game, this set summarizes all the gameplay mechanics of PvZ. These levels span across very easy and more difficult levels, with varied gameplay mechanics and various types and quantities of loadouts. For our generalization experiments, we train our agent in subsets of the original set of $40$ levels. We train the generalization agents with different amounts of levels: 5, 10, and 20. These subsets are randomly sampled from the original set. As we will see in Section~\ref{sec:generalization}, we then test the generalization agents in the full set of $40$ levels.

As mentioned in Section~\ref{sec:intro}, PvZ, as well as most tower defense games, includes a difficulty system. This system is used to keep the player engaged with the game: for more advanced players, the level should be more difficult than for less experienced ones. For this reason, in each level, we can set a level of difficulty. For our experiments, we train all of our agents for all levels with a fixed difficulty of 100K, which is generally high and makes most of the levels beatable only using mindful strategies. The value of the difficulty ultimately affects the composition, strength and health of the enemies in each wave. As we will see in Section~\ref{sec:difficulty_robustness}, to test the robustness of our agent, we vary the level of difficulty during testing between 0 and 200K, with the latter representing an almost impossible level to complete.

\subsection{Experiments and Baselines}
For this evaluation, we are mainly interested in 4 research questions: 
\begin{itemize}
    \item Is our proposed approach able to solve more levels than baselines?
    \item Is our proposed approach more robust to increasing game difficulty?
    \item Can our proposed approach generalize to unseen levels during training and still outperform baselines?
\end{itemize}

We will answer these questions later in Section~\ref{sec:results}. For comparison, two baselines were considered.
\begin{itemize}
    \item \textbf{Random}: we replace the priority calculator in HAI with a random number generator, sampling strategies following action masking as mentioned in Sections \ref{sec:action_mask};
    \item \textbf{HAI}: the full heuristic AI that consists of a hand-crafted priority calculator and low-level strategy executors. More details in Section~\ref{sec:hai}.
\end{itemize}

\begin{table*}[]
\centering
\caption{Performance of agents trained on subsets of the main levelset of 40 levels vs. HAI. The agents were trained on $N$ randomly sampled levels where $N \in \{5,10,20\}$. 
Values in the HRL\textsuperscript{\textdagger} column are identical to the ones in Table \ref{tab:main_table}, where one model was trained for each level respectively.
The final row presents the mean over all success rates and the sum of the episodic rewards.}
\label{tab:gen_table}
\scalebox{0.65}{
\begin{tabular}{c|cc|cc|cc|cc|cc}
      & \multicolumn{2}{c|}{HAI} & \multicolumn{2}{c|}{HRL\textsuperscript{\textdagger}}       & \multicolumn{2}{c|}{HRL ($N = 5$)}           & \multicolumn{2}{c|}{ HRL ($N = 10$)} &  \multicolumn{2}{c}{HRL ($N = 20$)}          \\
Level & Success Rate          & Reward  & Success Rate          & Reward                & Success Rate & Reward          & Success Rate   & Reward     & Success Rate & Reward \\ \hline
1 & \multicolumn{1}{c|}{$\phantom{0}14.00 \pm \phantom{00}3.90$} & \multicolumn{1}{c|}{$\mathbf{\phantom{0}11.56 \pm \phantom{00}0.82}$} & \multicolumn{1}{c|}{$\mathbf{\phantom{0}29.80 \pm \phantom{0}15.66}$} & \multicolumn{1}{c|}{$\phantom{0}11.40 \pm \phantom{00}5.23$} & \multicolumn{1}{c|}{$\phantom{00}7.60 \pm \phantom{00}2.65$} & \multicolumn{1}{c|}{$\phantom{00}7.93 \pm \phantom{00}2.26$} & \multicolumn{1}{c|}{$\phantom{0}14.60 \pm \phantom{00}7.55$} & \multicolumn{1}{c|}{$\phantom{00}6.54 \pm \phantom{00}5.50$} & \multicolumn{1}{c|}{$\phantom{00}6.60 \pm \phantom{00}4.59$} & \multicolumn{1}{c}{$\phantom{00}7.17 \pm \phantom{00}2.73$}\\
2 & \multicolumn{1}{c|}{$\phantom{0}45.80 \pm \phantom{00}2.40$} & \multicolumn{1}{c|}{$\phantom{00}8.41 \pm \phantom{00}0.48$} & \multicolumn{1}{c|}{$\mathbf{\phantom{0}67.00 \pm \phantom{00}9.61}$} & \multicolumn{1}{c|}{$\mathbf{\phantom{0}10.72 \pm \phantom{00}1.49}$} & \multicolumn{1}{c|}{$\phantom{0}56.40 \pm \phantom{0}10.48$} & \multicolumn{1}{c|}{$\phantom{0}10.22 \pm \phantom{00}1.47$} & \multicolumn{1}{c|}{$\phantom{0}43.60 \pm \phantom{0}24.98$} & \multicolumn{1}{c|}{$\phantom{00}6.05 \pm \phantom{00}6.85$} & \multicolumn{1}{c|}{$\phantom{0}32.20 \pm \phantom{0}15.45$} & \multicolumn{1}{c}{$\phantom{00}6.62 \pm \phantom{00}2.32$}\\
\end{tabular}
}
\end{table*}

\section{Results}
\label{sec:results}
In this section, experimental results are reported. Each experiment was repeated using different seeds for $5$ times over which the reported results are averaged and presented with mean and standard deviation. In all the figures and tables, our hybrid approach is identified as \textit{HRL}.

\subsection{General Performance}
\label{sec:general_performance}
In the first experiment, we train one HRL agent for each of the 40 levels in the training suite and compared these results against HAI and a random agent. Note that these levels are not in order of complexity, meaning that level 40 is not necessarily more complex than level 1. The results are reported in Table~\ref{tab:main_table}. Each of these agents was trained for 10K episodes at a fixed level of difficulty of 100K -- that is generally high for each level. As the table indicates, our approach achieves a generally higher success rate compared to the random agent and especially HAI. For the reasons detailed in Section~\ref{sec:hai}, the static strategy selection system is not optimal for playing all levels in this game. As the table shows, some levels are quite challenging to complete (e.g., levels 37, 38, and 39), with none of the approaches being able to solve them. However, in these levels our hybrid approach attains a higher reward compared to the others, meaning that it can eliminate more zombies and survive longer. Interestingly, there are levels (e.g., levels 10, 18, and 25) where even a random agent can achieve a 100\% success rate. In some instances (e.g., level 18), our agent attains a slightly lower success rate (99.80\%). Notably, there are some levels (e.g., levels 9, 21, and 26) where HAI performs significantly worse than a random agent, while this never occurs for HRL. Nonetheless, even though our hybrid approach achieves a higher overall success rate, we observe that HAI slightly outperforms HRL in some levels (e.g. level 7, 31, and 34). In total, our hybrid approach achieves a success rate of 57.12\%, compared to 47.95\% of HAI and 39.29\% of a random agent. At the same time, our agent achieves a much higher reward compared to the baselines: a total of 419.92 compared to 64.88 of HAI and 32.57 of the random agent.  Moreover, our agent has a higher total number of timesteps, meaning that it survives longer than the baselines. Even if there is not always a correlation between the number of steps and success rate, this value combined with the reward indicates that our agent survives longer \textit{and} kills more zombies, two important characteristics to win the game. In case of HAI, there is one particular level (level 23) where this baseline achieves a very low reward that decreases the total value. However, even if we remove this outlier, our approach still achieves a higher reward.

\begin{figure}
\centering
\begin{tabular}{cc}
    \centering
    \includegraphics[width=0.48\columnwidth]{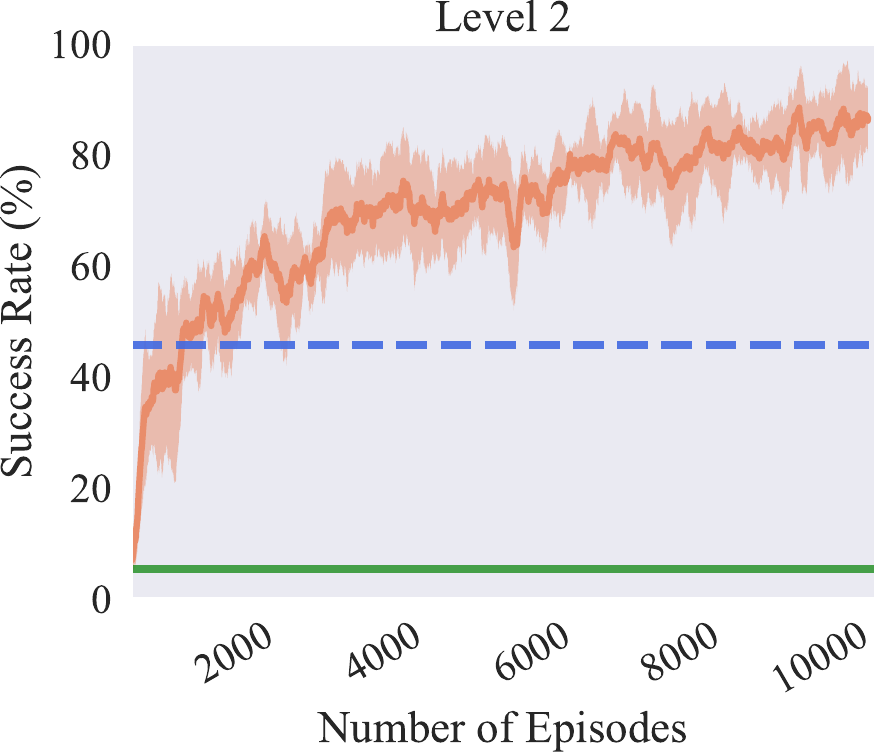} &
    \includegraphics[width=0.48\columnwidth]{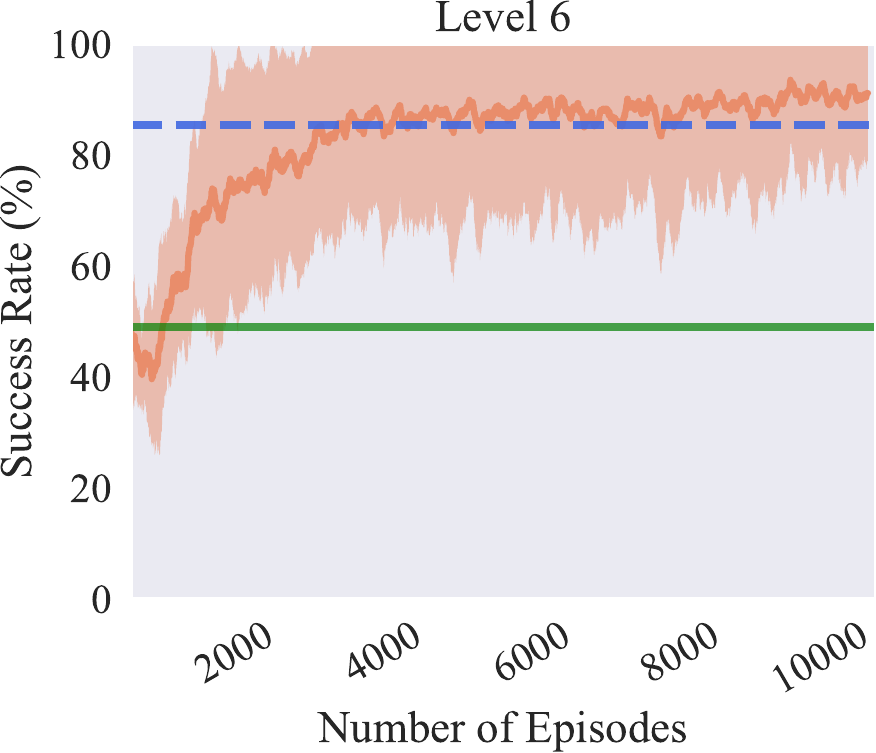} \\
    (a) &  (b) \\
    \includegraphics[width=0.48\columnwidth]{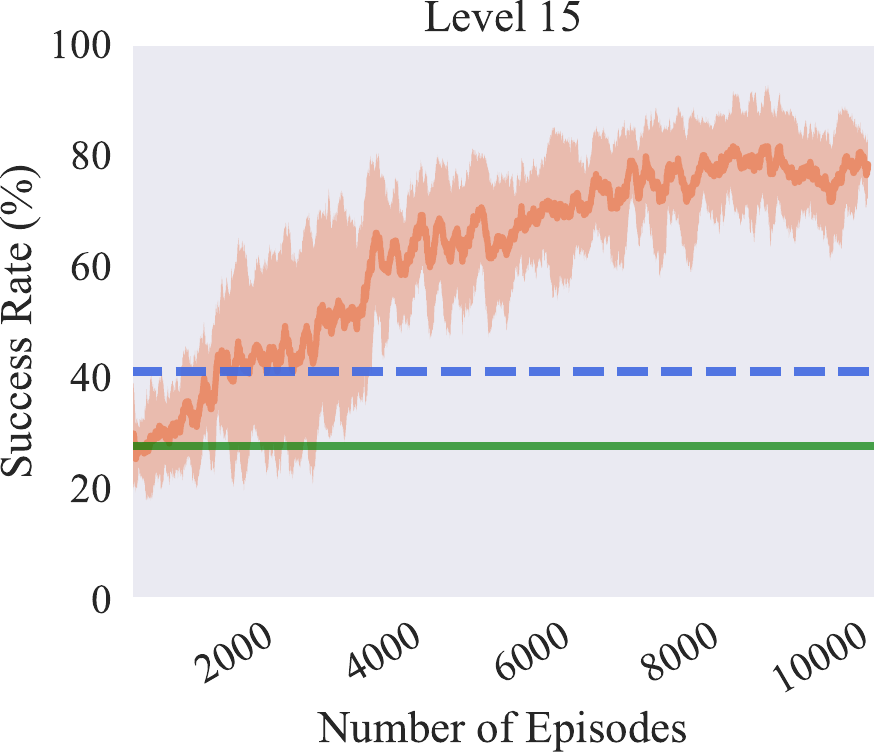} &
    \includegraphics[width=0.48\columnwidth]{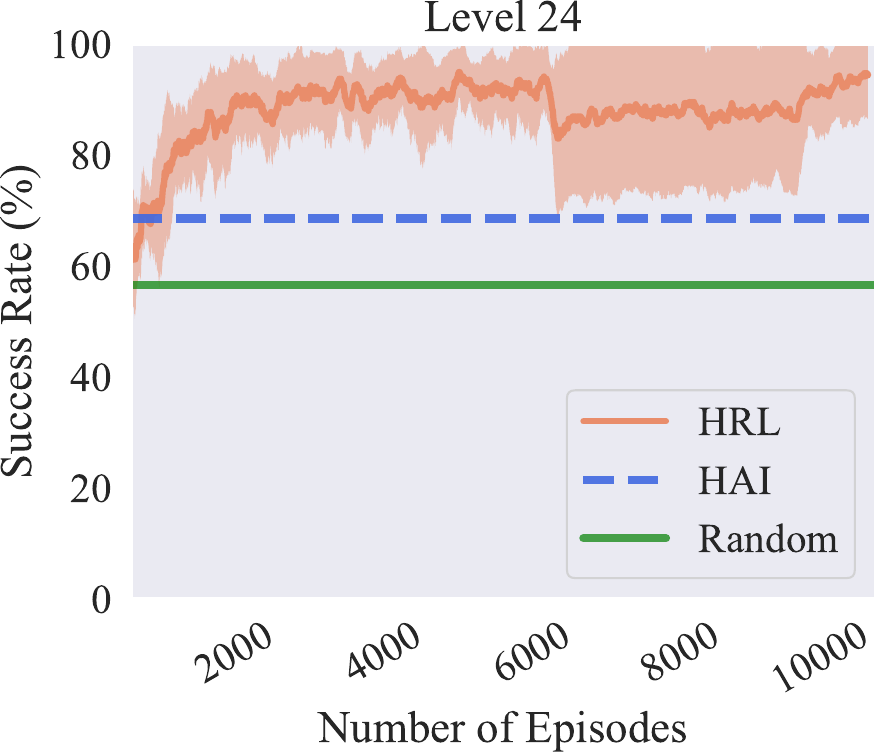} \\
    (c) & (d)\\
    \end{tabular}
\caption{Training progression of the HRL agent (orange) over four levels, compared to the mean success rate of HAI (blue) and a random agent (green) over 100 episodes in each level respectively. In the displayed levels, the HRL agent learns to outperform HAI in approximately 400 to 1500 episodes. Both the HRL agent and HAI consistently outperform the random agent. The gathered statistics are averaged over $5$ different seeds.}
\label{fig:training_progression}
\end{figure}

In Figure~\ref{fig:training_progression}, we show the training progression of the agent across four levels: levels 2, 6, 15, and 24. We compare the training progression with the success rates of HAI and the random agent. As the baselines are not trained, their values remain constant regardless of the training. From these plots we can see that our agent starts to outperform HAI in approximately 400 to 1500 training episodes.

\begin{figure}[t]
\centering
\begin{tabular}{cc}
    \centering
    \includegraphics[width=0.48\columnwidth]{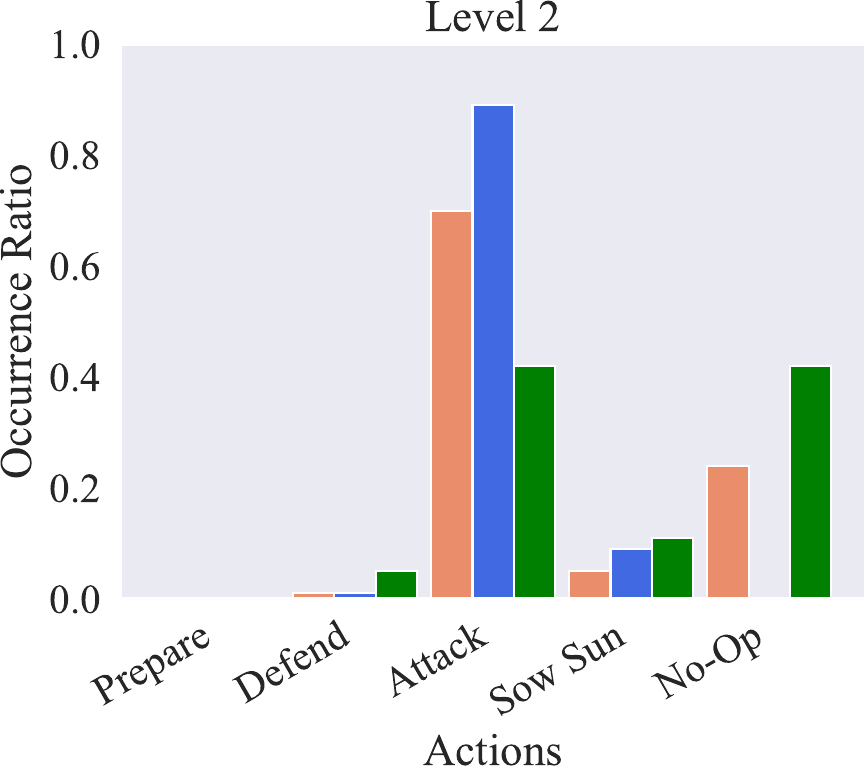} &
    \includegraphics[width=0.48\columnwidth]{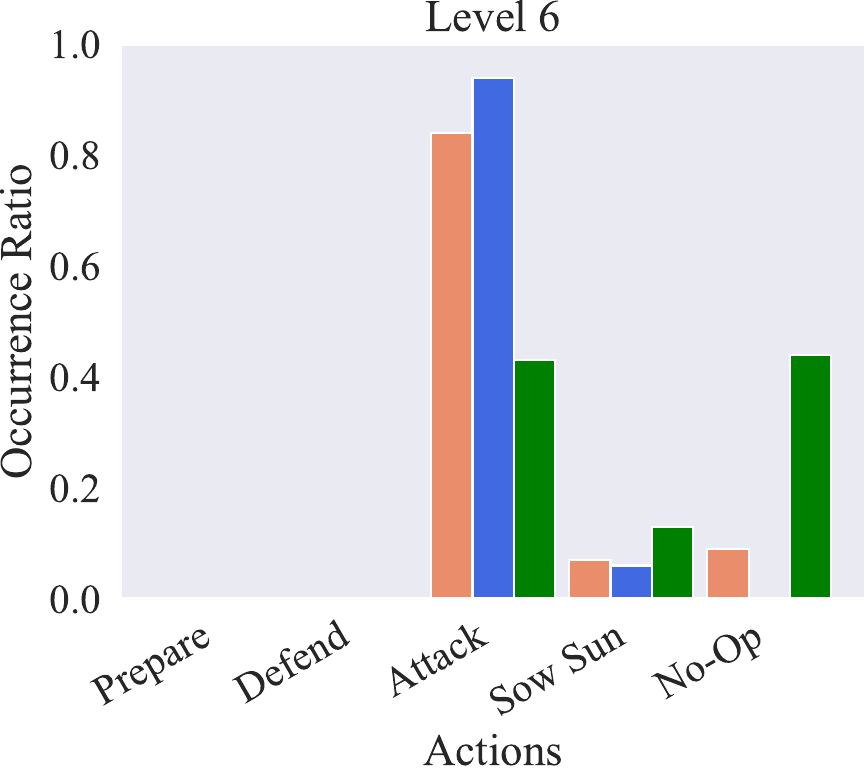} \\
    (a) &  (b) \\
    \includegraphics[width=0.48\columnwidth]{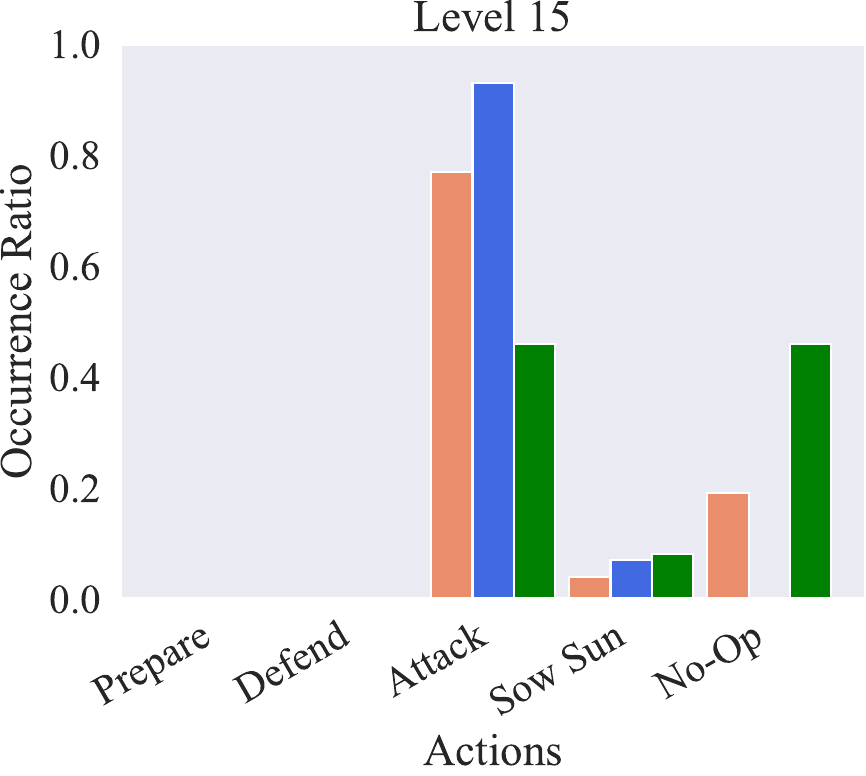} &
    \includegraphics[width=0.48\columnwidth]{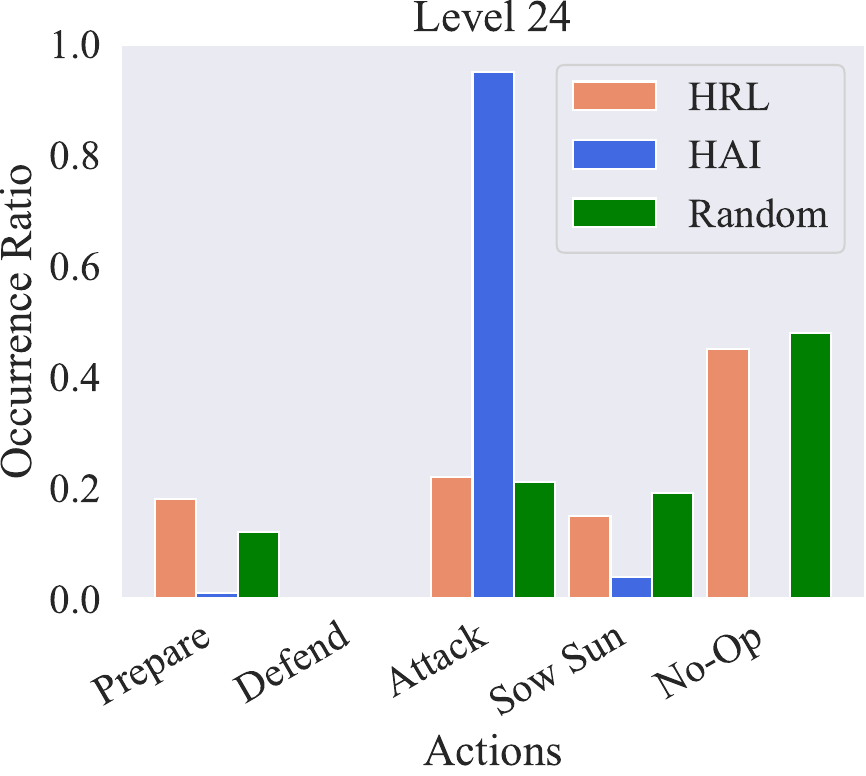} \\
    (c) & (d)\\
    \end{tabular}
\caption{Action distribution comparison between HAI, random and HRL agents over four levels representing the average, normalized occurrence of each action over 100 episodes repeated for $5$ different seeds. As evident from the plots, the HRL agent utilizes the actions differently than the other baselines from level to level, meaning the agent has learned to better leverage actions based on the dynamics of the level.}
\label{fig:action_distribution}
\end{figure}

It is interesting to compare the general behavior between HAI and our hybrid approach. Figure~\ref{fig:action_distribution} shows the action distributions for four example levels: levels 2, 6, 15, and 24. As illustrated by the plots, HAI exhibits a considerably more aggressive behavior in all the levels, with no significant differences between behaviors across different levels. In contrast, our hybrid approach demonstrates a more cautious behavior in all levels, but in varying ways: in levels 2 and 15, our hybrid approach waits for the opportune moment to deploy a defensive unit, while in levels 6 and 24, it plants additional sunflower units to gather more resources for deploying more effective offensive units.

\setlength{\tabcolsep}{1pt}

\subsection{Generalization Performance}
\label{sec:generalization}
In this experiment, we train a set of three agents: one trained in 5 levels, another in 10 levels, and the last one in 20 levels. All agents were trained for 10K episodes. The levels chosen for training these agents were randomly sampled from the original list of 40 levels. We repeat this experiment for 5 seeds, each time sampling a different subset of levels. In contrast to Section~\ref{sec:general_performance}, where we trained one agent for each level, these agents were exposed to multiple levels during training and then tested on unseen levels without retraining. Our goal is to evaluate the ability of our approach to generalize to unseen levels and determine the number of levels needed for training an agent that generalizes effectively. Table~\ref{tab:gen_table} presents the results of this experiment. As the table indicates, these agents are unable to generalize sufficiently to outperform the agents trained in Section~\ref{sec:general_performance}, and especially HAI. This could imply that each level requires a specific, non-transferable strategy. This observation reflects one of the main characteristics of tower defense games: each level is a puzzle that necessitates its own strategy, and players must fail before discovering the correct one. This is precisely what our agents trained in Section~\ref{sec:general_performance} do: they fail and learn for each level.

\subsection{Difficulty Robustness}
\label{sec:difficulty_robustness}
\begin{figure}
\centering
\begin{tabular}{cc}
    \centering
    \includegraphics[width=0.48\columnwidth]{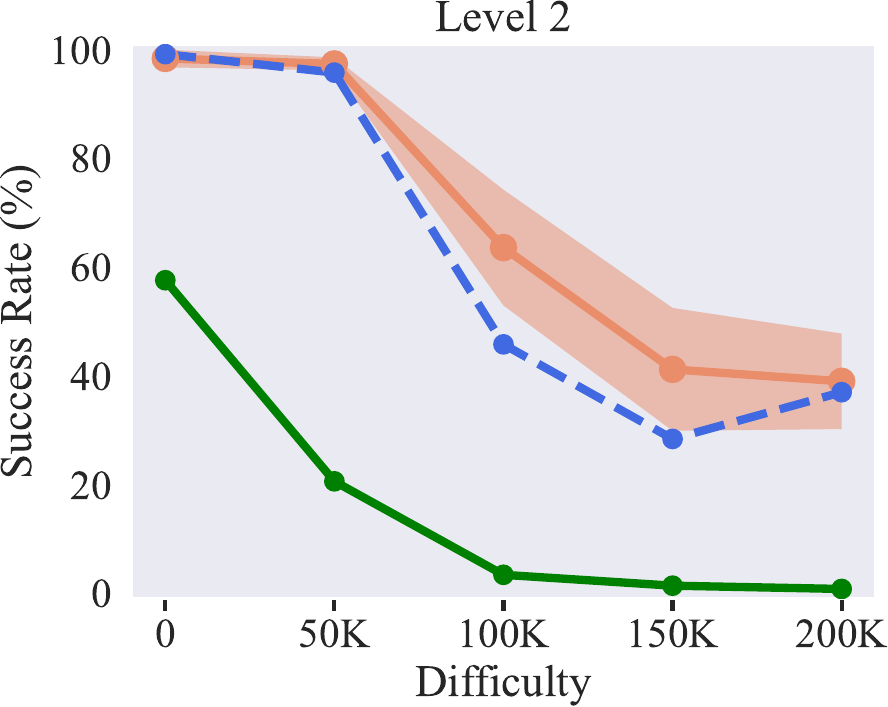} &
    \includegraphics[width=0.48\columnwidth]{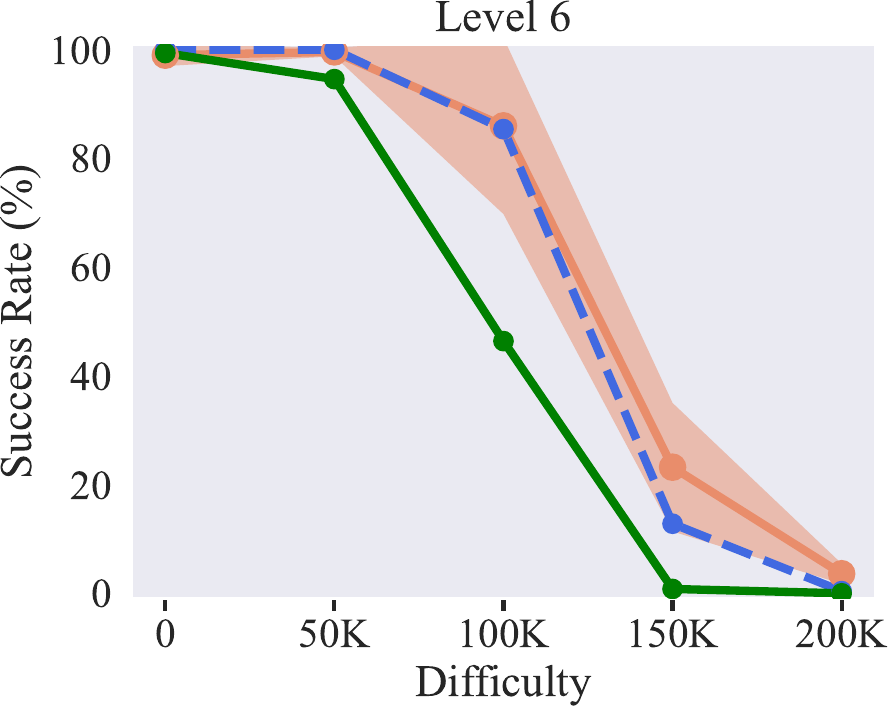} \\
    (a) &  (b) \\
    \includegraphics[width=0.48\columnwidth]{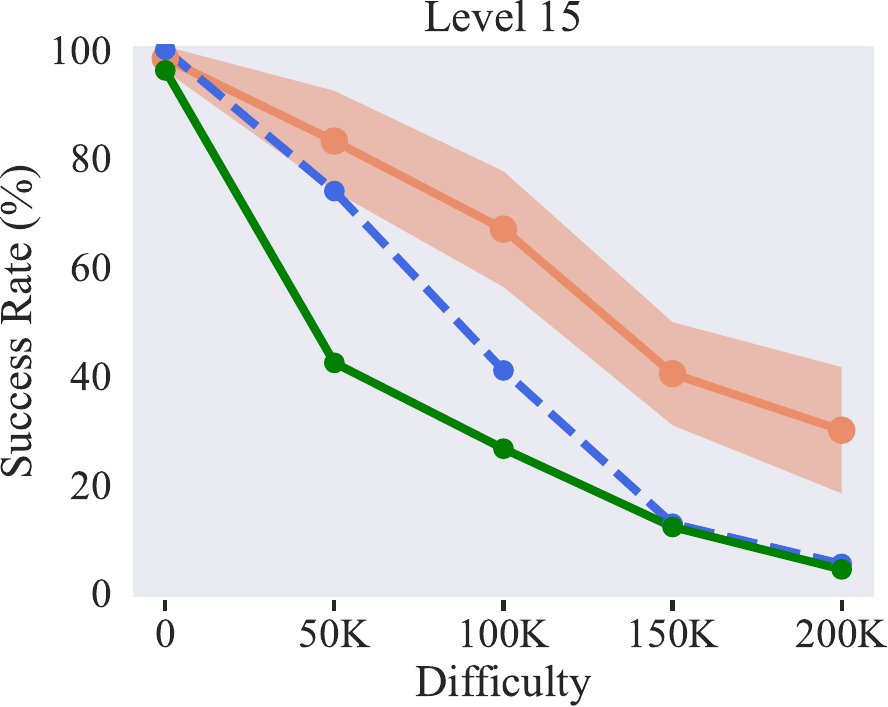} &
    \includegraphics[width=0.48\columnwidth]{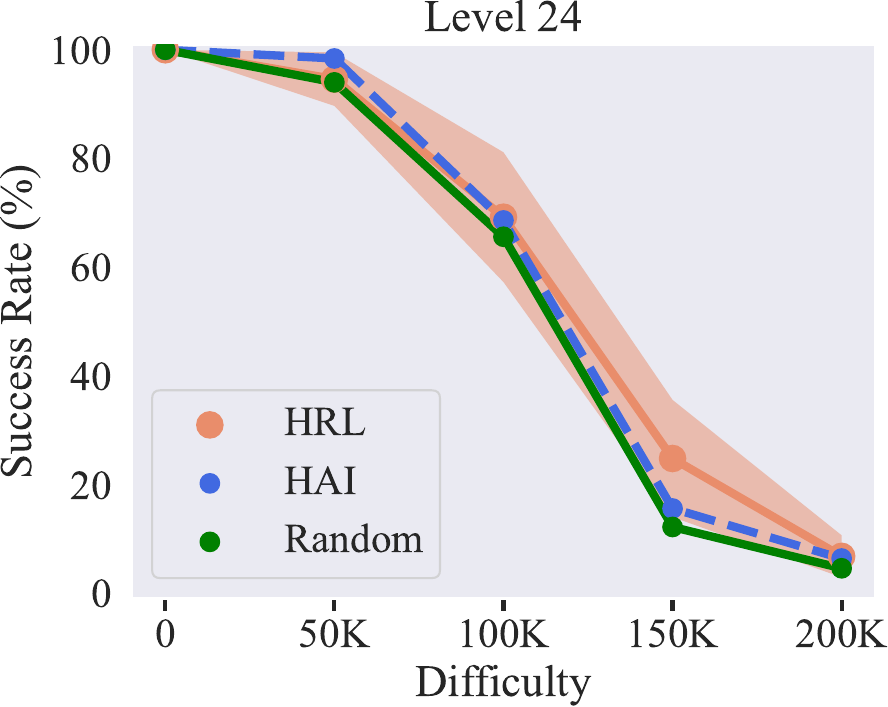} \\
    (c) & (d)\\
    \end{tabular}
\caption{Performance of HRL agent (orange) compared to HAI (blue) and a random agent (green) with increasing difficulty in 4 levels. Five difficulties were used, specifically [0, 50K, 100K, 150K, 200K]. More details about the difficulty can be found in Section~\ref{sec:levels_and_difficulty}. The general performance of the agent indicates the it successfully completes more levels than the baselines for each difficulty.  We collect statistics for the experiments for 100 levels and 5 different seeds. Note, in the experiments, the agents were trained with difficulty of 100K, the mid-point of the aforementioned difficulty set.}
\label{fig:increasing_difficulty}
\end{figure}

In this experiment, we use the agents trained in Section~\ref{sec:general_performance}. The goal of this experiment is to determine whether the performance of trained agents changes when the level of difficulty of a particular puzzle is altered, and how it compares to HAI. We conduct this experiment for four different levels -- level 2, 6, 15, and 24 -- and for five different difficulty levels -- [0, 50K, 100K, 150K, 200K]. We have chosen these levels as they are levels where all the baselines perform reasonably well at difficulty level 100K (see Table~\ref{tab:main_table}). Figure~\ref{fig:increasing_difficulty} displays the results. As evident from the figure, our hybrid RL solution is capable of maintaining higher performance even at higher difficulties compared to HAI. This demonstrates that our agent finds a better solution for a specific puzzle regardless of its difficulty level, in contrast to HAI.

\section{Conclusion and Discussion}
In this paper, we propose a hybrid approach that combines RL and heuristic AI, thus integrating the performance of an RL-trained agent with the scalability of hard-coded scripted AI. We tested our approach in the popular tower defense game, \textit{Plants vs. Zombies}. We found that leveraging existing expertise and knowledge of a classical AI system and augmenting its capabilities using a self-learning strategy prioritizer leads to improved performance over the original AI system. 

Our experiments show that this approach can approximate high-performing players, but also their adaptability, that could be leveraged to generate data useful for level testing and validation at scale. With little human intervention, we show that the approach can be applied to previously unseen levels simply through re-training. However, we also conclude that the nature of tower defense games, which are very puzzle-like in their nature, makes it hard to train a general solution for all levels. The recommendation is to rather train a model for each level to get consistent results.

Although our approach demonstrated promising results in the tested game, we believe there is still room for improvement. As we saw little evidence of generalization over previously unseen levels both for the HAI and the hybrid solution important future work here would be to create an algorithm that can handle levels/features/situations that are fundamentally different without the need of retraining. We also see that other game genres, which uses a HAI for low-level control but also have a strategic component that can be replaced by RL, can benefit from this proposed solution. For instance, we believe real-time strategy games would be of interest for future application of this approach. Even for team based action games with a capable low-level heuristic AI system for locomotion, navigation, and combat, this approach could be used to improve high-level strategy by choosing which low-level behavioral routine the bots should run.

\section*{Acknowledgments}
We would like to express our gratitude to Timur Solovev and Mónica Villanueva Aylagas for their valuable contributions, as well as Cory Deines, Rain Zhang and Geoff Schemmel from the \textit{PopCap} team for their extensive support.

\bibliographystyle{IEEEtranN}
{\footnotesize \bibliography{refs}}
\end{document}